\documentclass{article}

\usepackage[preprint, nonatbib]{neurips_2021}

\usepackage[utf8]{inputenc} 
\usepackage[T1]{fontenc}    
\usepackage{hyperref}       
\usepackage{url}            
\usepackage{booktabs}       
\usepackage{amsfonts}       
\usepackage{nicefrac}       
\usepackage{microtype}      
\usepackage[round, authoryear]{natbib}
\usepackage{tikz}
\usepackage{bm}
\usepackage{tabularx}
\usepackage{multirow, makecell}
\usepackage{graphicx}
\usepackage{caption}
\usepackage{subcaption}
\usepackage{comment}
\usepackage{color}
\usepackage[shortlabels]{enumitem}
\usepackage{wrapfig}

\hypersetup{
	colorlinks=true,
	linkcolor=blue,
	citecolor=black,
	urlcolor=blue
}

\graphicspath{{images/}}

\usetikzlibrary{shapes,arrows,arrows.meta,positioning,fit,backgrounds,matrix, decorations.pathreplacing, calligraphy}

\tikzstyle{arrow} = [very thick,->,>=stealth]
\tikzstyle{var} = [ellipse, draw=black, minimum width=2.0cm, minimum height = 1.0cm]
\tikzstyle{process} = [rectangle, rounded corners, draw=black, minimum width=3cm, minimum height = 1.5cm, align=center]
\tikzstyle{arrow2} = [>={latex[width=6mm,length=5mm]}, ->, line width=3pt]
\tikzset{
	position/.style args={#1:#2 from #3}{
		at=(#3.#1), anchor=#1+180, shift=(#1:#2)
	}
}

\title{PyHard: a novel tool for generating hardness embeddings  to support data-centric analysis}

\author{%
  Pedro Yuri Arbs Paiva \\
  Instituto Tecnológico de Aeronáutica (ITA) \\
  Praça Marechal Eduardo Gomes, 50 \\
  São José dos Campos, SP, Brazil \\
  \texttt{paiva@ita.br} \\
  \And
  Kate Smith-Miles \\
  School of Mathematics and Statistics \\
  The University of Melbourne \\
  Parkville VIC 3010, Australia \\
  \texttt{smith-miles@unimelb.edu.au}
  \And
  Maria G. Valeriano
  \\
  Instituto Tecnológico de Aeronáutica (ITA) \\
  Praça Marechal Eduardo Gomes, 50 \\
  São José dos Campos, SP, Brazil \\
  \texttt{maria.valeriano@ita.br} 
  \And
  Ana Carolina Lorena \\
  Instituto Tecnológico de Aeronáutica (ITA) \\
  Praça Marechal Eduardo Gomes, 50 \\
  São José dos Campos, SP, Brazil \\
  \texttt{aclorena@ita.br} \\
}

\begin{document}

\maketitle

\begin{abstract} 
For building successful Machine Learning (ML) systems, it is imperative to have high quality data and well tuned learning models. But how can one assess the quality of a given dataset? And how can the strengths and weaknesses of a model on a dataset be revealed? Our new tool PyHard employs a methodology known as Instance Space Analysis (ISA) to produce a hardness embedding of a dataset relating the predictive performance of multiple ML models to estimated instance hardness meta-features. This space is built so that observations are distributed linearly regarding how hard they are to classify. The user can visually interact with this embedding in multiple ways and obtain useful insights about data and algorithmic performance along the individual observations of the dataset. We show in a COVID prognosis dataset how this analysis supported the identification of pockets of hard observations that challenge ML models and are therefore worth closer inspection, and the delineation of regions of strengths and weaknesses of ML models. 
\end{abstract}

\section{Introduction}

A recent Meta-Learning (MtL) \citep{Vilalta:mtl-survey,vanschoren2019meta} methodology known as Instance Space Analysis (ISA) has demonstrated success in gaining insights through the visual analysis of the relationships between the performance of ML classification techniques and data meta-features across multiple datasets \citep{Munoz:is}. Using ISA, the performance of multiple ML classification techniques can be predicted and understood in terms of a set of meta-features describing popular classification datasets from public repositories. A 2-D projection relating meta-features values and algorithmic performance is then created, presenting linear trends that reveal pockets of hard and easy datasets, that is, datasets that are harder or easier for the algorithms to classify, revealing the strengths and weaknesses of each classifer.

We adapt here such a framework for a fine-grained analysis of a single  classification dataset, rather than a collection of datasets, so that we can ontain similar insights about the difficulties of individual observations within a dataset. Given a dataset of interest, multiple measures of the hardness level of its observations are extracted and related to the algorithmic performance of popular ML models. As a result, a 2-D hardness embedding of the original dataset is obtained, which can be scrutinized to inspect data quality and understand classifier behaviors. We present herein \textit{PyHard}, an analytical tool designed to assist the understanding of the difficulty level of the individual observations in a dataset, and also to trace regions where each algorithm is expected to perform well. We show in a case study with a dataset for COVID-19 prognosis how this tool supported the identification of problematic observations and the understanding of the behavior of different classifiers across the dataset.

\section{Instance Space Analysis for a single dataset}\label{sec:ISA}

As originally proposed, the Instance Space Analysis (ISA) builds upon the Algorithm Selection Problem (ASP) \citep{Rice:asp, Smith-Miles:asp}, whose objective is to automate the process of selecting good candidate algorithms and their hyperparameters for solving new problems, based on knowledge gathered from similar problems they solved in the past. The ISA framework goes further and extends the ASP analysis to give insights into why some instances are harder to solve than others, combining the information of meta-features and algorithm performance in a new embedding that can be visually inspected. To this end, an optimization problem is solved to find the mapping from the high dimensional meta-features space into a 2-D space, such that the distribution of algorithm performance metrics and meta-feature values across instances in the 2-D space displays as much of a linear trend as possible to assist the interpretation of hardness directions. Footprints of algorithm performance representing regions where each algorithm performs well can also be delineated, allowing to characterize their domains of competence. Objective measures of algorithmic performance along the IS can be extracted from the footprints too, such as area, density and purity. Details on the construction of the instance space and calculation of algorithm  footprints can be found at \citep{Munoz:is,munoz2021regression}.

\begin{wrapfigure}{r}{0.7\textwidth}
    \centering
    \resizebox{0.7\textwidth}{!}{%
        \centering
        \begin{tikzpicture}[node distance=5cm]
            \tikzstyle{every node}=[font=\Large]
            \node (data) [process, fill=gray!5] {Original \\ dataset $D$};
            \node (measure) [process, fill=gray!5, above right =of data, yshift=-3cm, xshift=-10mm] {$f(\mathbf{x}_i) \in \mathcal{F}$\\[2pt]Hardness\\ measures};
            \node (algo) [process, fill=gray!5, below right =of data, yshift=3cm, xshift=-10mm] {$\alpha \in \mathcal{A}$\\Algorithm\\ space};
            \node (perf) [process, fill=gray!5, right =20mm of algo] {$y \in \mathcal{Y}$\\[2pt] Performance\\ space};
            \node (fs) [process, fill=gray!5] at (measure -| perf) {$fs(f(\mathbf{x}_i), y)$\\[2pt] Feature\\ selection};
            \node (meta) [process, right=110mm of data.east, fill=gray!5] {$\mathbf{\mu} \in \mathcal{M}$\\[2pt]Metadata};
            \node (is) [process, fill=gray!5, right =15mm of meta] {$(z_1, z_2) \in \mathbb{R}^2$\\[2pt] Instance space};
            
            \coordinate[right=10mm of data.east] (aux1);
            
            \draw [thick, ->,>=stealth] (data.east) -- (aux1) |- node[pos=0.75, above, align=center] {$f(\cdot)$} (measure.west);
            \draw [thick, ->,>=stealth] (aux1) |- (algo.west);
            \draw [thick, ->,>=stealth] (algo) -- node[pos=0.5, above, align=center] {$y(\alpha, \bm{x}_i)$} (perf);
            \draw [thick, ->,>=stealth] (measure) -- (fs);
            \draw [thick, ->,>=stealth] (perf.north) -- (fs.south);
            \draw [thick, ->,>=stealth] (fs) -| (meta);
            \draw [thick, ->,>=stealth] (perf) -| (meta);
            \draw [thick, ->,>=stealth] (meta) -- node[pos=0.5, above, align=center] {$ISA$} (is);
        \end{tikzpicture}
    }
    \caption{ISA framework for a single dataset.}
    \label{fig:ih}
\end{wrapfigure}
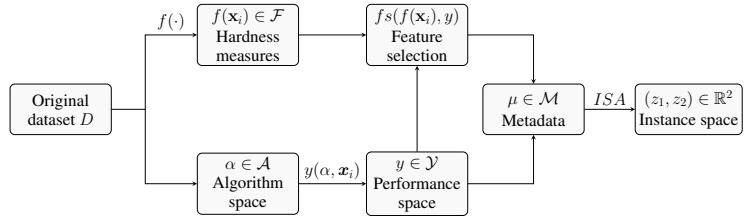

Here we recast the framework for the analysis of a single classification dataset, as outlined in Figure \ref{fig:ih}. Given a dataset 
$\mathcal D$ with $n$ labeled observations $\mathbf x_i$, a set of \textit{hardness measures} (HM) $\mathcal F$ assessing their difficulty (aka hardness level) is extracted. A pool of ML classifiers of distinct biases are applied to the dataset and compose the algorithm space $\mathcal{A}$, namely:  Bagging, Gradient Boosting, Support Vector Machine (both linear and RBF kernels), Logistic Regression, Multilayer Perceptron and Random Forest (but alternative algorithms can be added). The performance space $\mathcal{Y}$ records the performance obtained by each algorithm in $\mathcal A$ for each instance $\mathbf x_i \in D$. A meta-dataset $\mathcal M$ is built by joining a subset of $\mathcal F$ and the set $\mathcal{Y}$. The former subset contains meta-features which relates more to the performance space $\mathcal Y$, which are screened in a feature selection step. ISA is then applied for obtaining the 2-D hardness embedding (instance space). Some of the most specific components of our framework are described next.

\paragraph{Hardness measures.}
One important aspect when performing ISA is using a set of informative meta-features able to reveal the capabilities of the algorithms and the level of difficulty each individual instance poses. Here we revisit the definition of \textit{Instance Hardness} (IH)  proposed by \citet{Smith:hardness} as a property that indicates the likelihood that an instance will be misclassified. IH is measured by Equation \ref{eq:IH}, where $p\big(c_i | \mathbf x_i, \alpha_j\big)$ gives the probability a learning model $\alpha_j$ induced from $D$ assigns $\mathbf x_i$ to its expected class $c_i$. According to this formulation, observations that are frequently misclassified by a pool $\mathcal A$ of diverse learning algorithms are considered hard, whilst easy instances are likely to be correctly classified by any of the considered algorithms.
\begin{equation}\label{eq:IH}
     IH_{\mathcal{A}}\big(  \mathbf x_i, c_i  \big) = 1 - \frac{1}{|\mathcal{A}|} \sum_{j=1}^{|\mathcal{A}|} p\big(c_i | \mathbf x_i, \alpha_j\big)
\end{equation}

\begin{table}[ht]
    \centering
    \caption{Hardness measures employed as meta-features in this work.}
    \resizebox{0.65\columnwidth}{!}{%
    {
    \begin{tabular}{llccl}
         \hline
         \bf Measure & \bf Acron. & \bf Min & \bf Max & \bf Reference  \\
         \hline
         k-Disagreeing Neighbors& $kDN$ & 0 & 1 & \citep{Smith:hardness} \\
         Disjunct Class Percentage& $DCP$  & 0 & 1 & \citep{Smith:hardness} \\
         Tree Depth (pruned)& $TD_P$  & 0 & 1 & \citep{Smith:hardness} \\
         Tree Depth (unpruned)& $TD_U$  & 0 & 1 & \citep{Smith:hardness} \\
         Class Likelihood& $CL$  & 0 & 1 & \citep{Smith:hardness} \\
         Class Likelihood Difference& $CLD$  & 0 & 1 & \citep{Smith:hardness} \\
         Frac. features in overlapping areas & $F1$ & 0 & 1 & \citep{Arruda:ih} \\
         Frac. nearby instances of different class & $N1$& 0 & 1 & \citep{Arruda:ih} \\
         Ratio of intra-extra class distances & $N2$& 0 & $\approx 1$ & \citep{Arruda:ih} \\
         Local set cardinality & $LSC$& 0 & 1 & \citep{Arruda:ih} \\
         Local set radius & $LSR$& 0 & 1 & \citep{Arruda:ih} \\
         Usefulness & $U$& $\approx 0$ & $1$ & \citep{Arruda:ih} \\
         Harmfulness & $H$& 0 & $\approx 1$ & \citep{Arruda:ih} \\
         \hline
    \end{tabular}}%
    }
    \label{tab:HM}
\end{table}

\citet{Smith:hardness} also define a set of \textit{hardness measures} (HM) intended to explain why some instances are often misclassified. These are the measures employed as meta-features in $\mathcal F$. Table \ref{tab:HM} summarizes the HM employed in this work, with their names, acronyms, minimum and maximum values achievable and references from where they are extracted. We introduced modifications into some of the measures in order to limit and standardize their values. Consequently, for all measures higher values are registered for instances that are harder to classify.

\paragraph{Algorithm performance assessment.}
Given a dataset $D$, we first split it into five folds according to the cross-validation (CV) strategy, such that each instance belongs to one of the test sets.  For each instance, we measure the \textit{log-loss error rate} achieved in the predictions, which takes into account the probabilities associated to each class. Furthermore, a hyper-parameter optimization step was added, acting as an inner loop for each of the training sets of the outer CV. Within this inner loop, a candidate set of parameters is evaluated through cross-validation upon the training data from the outer loop.

\paragraph{Feature selection.}
According to \citet{Munoz:is}, it is advisable to maintain just the most informative meta-features in the meta-dataset $\mathcal M$ before the IS projection is generated. We performed a supervised meta-feature selection in $\mathcal M$, based on the continuous response value $y(\alpha_k, \mathbf {x}_i)$, that is, the log-loss performance of the classifiers for the instances in $\mathcal D$. Since there are seven classification algorithms in the pool, a ranking of meta-features for each one of them is obtained. Next, a rank aggregation method is employed to merge these subsets, as suggested in \citep{prati2012combining}.

\paragraph{IS representation and footprints.}
PyHard embeds a Python package named PyISpace implementing the main ISA functionalities, including obtaining the projections and the algorithms' footprints. In order to assist the visual interpretation of the 2-D projections, we introduce a rotation step in order to ensure that hard instances are always placed towards the upper left corner of the IS, whilst the easier instances are placed towards the bottom right corner of the space. We also introduce an \textit{instance easiness footprint}, which are regions of the IS which encompass observations with IH values below a threshold of 0.4. These are the easiest instances of the dataset for most of the algorithms.

\section{Case study: inspecting a COVID prognosis dataset} \label{sec:exp}

Here we present an analysis of a dataset containing anonymized data from citizens positively diagnosed for COVID-19, collected from March 1st, 2020 to April 15th, 2021\footnote{The data comes from a partnership with the health department of the city of São José dos Campos - SP, Brazil. Part of such data was formatted for some predictive analysis to support public health decision making.}. The dataset involves predicting whether an individual will require future hospitalization or not taking as input the following attributes routinely supplied during COVID scanning: age, sex, initial symptoms (fever, cough, sore throat, dyspnea, respiratory distress, low saturation, diarrhea, vomit and other symptoms) and comorbidities (chronic cardiovascular disease, immunodeficiency-immunodepression, chronic kidney disease, diabetes mellitus, obesity, chronic respiratory diseases and other risks). The dataset has data from 5,156 citizens, half of which were hospitalized.

\begin{figure}[ht]
    \begin{subfigure}[t]{0.30\textwidth}
    \centering
        \includegraphics[width=\textwidth, trim=0 2.3cm 0 0,clip]{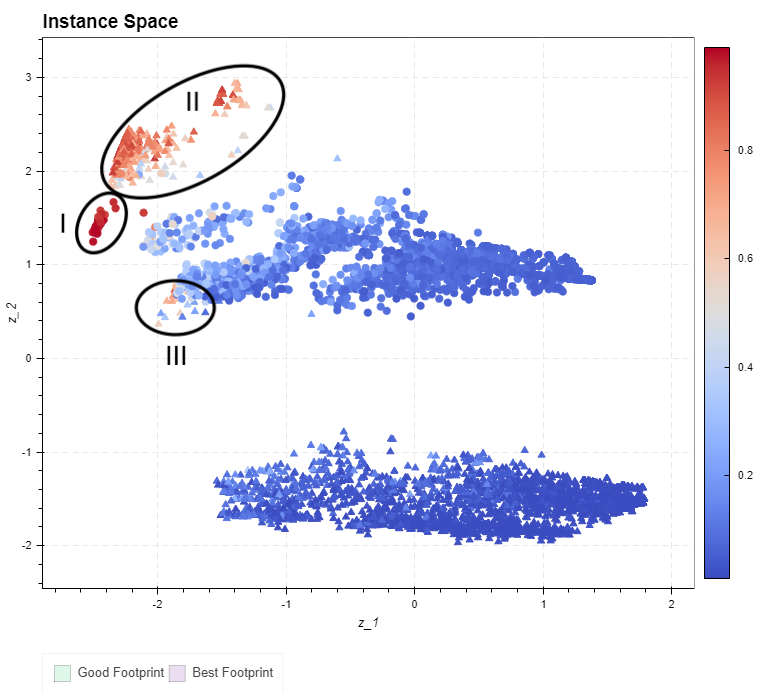}
        \caption{Colored by $IH$ values. }
    \end{subfigure}
    \hfill%
    \begin{subfigure}[t]{0.30\textwidth}
    \centering
        \includegraphics[width=\textwidth, trim=0 2.3cm 0 0,clip]{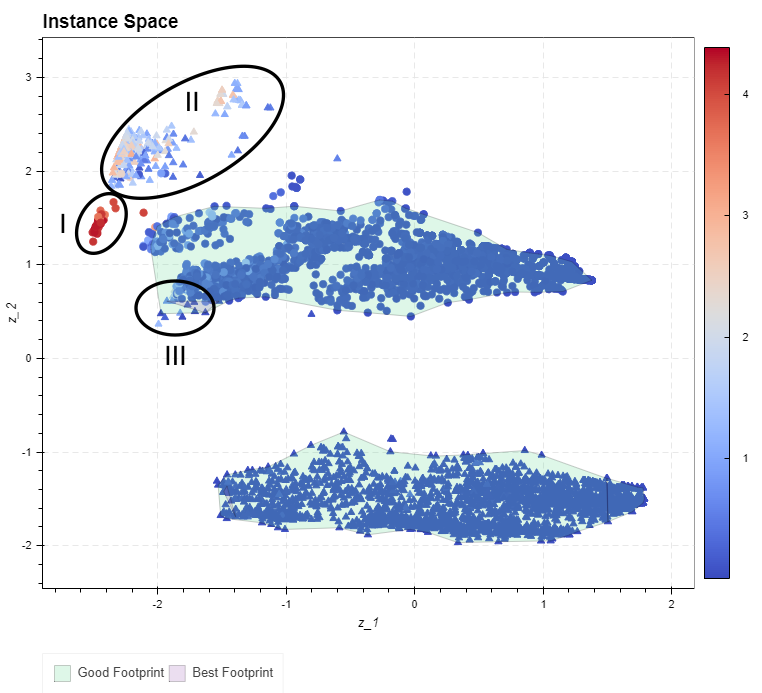}
        \caption{Colored by MLP log-loss.}
    \end{subfigure}
    \hfill%
    \begin{subfigure}[t]{0.38\textwidth}
    \centering
        \includegraphics[width=0.75\textwidth,height=0.62\textwidth]{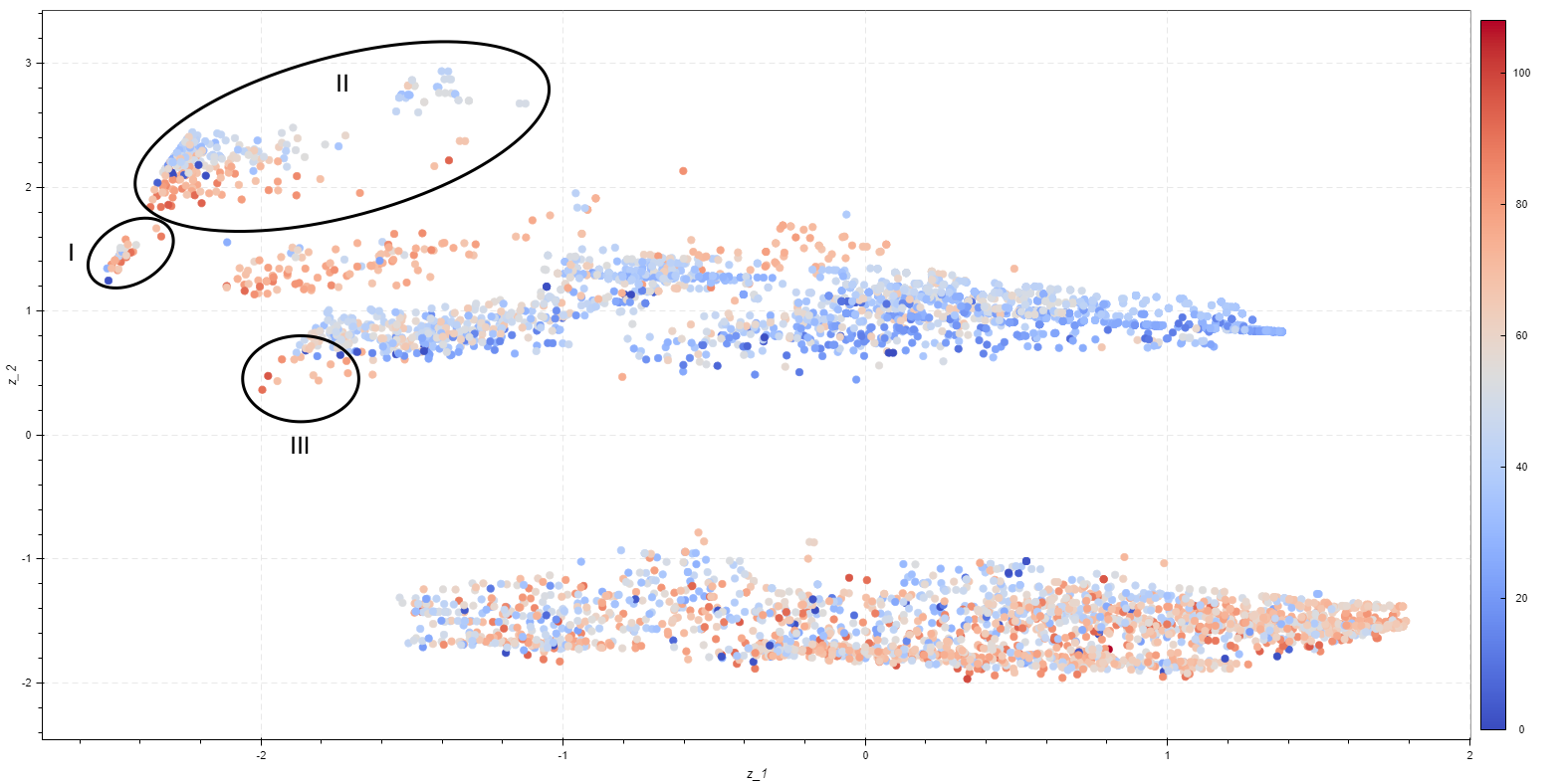}
        \caption{Colored by age input feature value.}
    \end{subfigure}
    \hfill%
    \caption{ISA for the hospitalization dataset.}
    \label{fig:hosp}
\end{figure}

 Our objective is to analyze the hardness profile of this dataset and to extract some insights from the visualization and interaction with its IS\footnote{The complete experiment with anonymizations can be found at: \url{https://gitlab.com/ita-ml/pyhard/-/tree/master/experiments/hospitalization}. Instructions can be found in `experiments' folder.}. Figure \ref{fig:hosp} presents the ISA of the hereafter called hospitalization dataset. Each point corresponds to an observation of the dataset, that is, a confirmed COVID case. In Figure \ref{fig:hosp}a, the observations are colored according to their IH values, with harder observations colored in red and easier observations  colored in blue, hospitalized individuals plotted as triangles and non-hospitalized observations represented by circles. Most of the observations in this dataset are easy to classify correctly by the algorithm portfolio and the instance easiness footprint has a large area. 

There is a large group of hospitalized citizens with an easier profile placed in the bottom of the IS. Non-hospitalized observations had mostly an intermediate hardness level, with low to medium IH. But some particular clusters of observations hard to classify, highlighted in circles numbered as I, II and III have caught our attention. Group I contains observations from the non-hospitalized class, whilst groups II and III have mostly instances from the hospitalized class. The observations from these groups have: a low likelihood of belonging to their classes (measured by $CL$ meta-feature); are placed in disjuncts with elements which do not share their labels (measured by $DCP$); are  close to elements from the opposite class (measured by $N1$); and have a high proportion of nearest neighbors with labels which differ from their own (measured by $kDN$). Overall, this indicates these groups probably contain noisy or outlier instances with input characteristics similar to that of other class. Plotting the values of the raw input attributes along the IS allowed us to observe some interesting features. Figure \ref{fig:hosp}c shows one of them: age (red colors are attributed to elderly and youngsters are colored in blue), a well known feature with influences on COVID severity, impacting hospitalization. Indeed, the easiest cases of non-hospitalized citizens tend to be younger. But some observations from groups I and II have conflicting age patterns to what is commonly expected. Contradicting patterns are also observed for other input features along the observations from groups I, II and III.  Inspecting the original records from these groups of citizens, we were able to confirm that group I has either individuals wrongly labeled as non-hospitalized and some atypical cases, such as people seeking for hospitalization too late. Group III has individuals with few and mild symptoms who were hospitalized and it is possible that their forms have missing information on some of the symptoms. Group I has also atypical cases which may contain missing information.

Concerning algorithmic performance, Figure \ref{fig:hosp}b presents the ISA, colored according to the log-loss performance of the MLP classifier which attained the largest footprint area (shown within a green polygon). The MLP showed a good predictive performance for most of the instances except from those of the groups I and II. The PyHard tool allows to save and to inspect the characteristics of the instances from a selected footprint for gain further insights into an algorithm's strengths and weaknesses, and the explanations for such based on the characteristics of these instances.

\section{Conclusion}\label{sec:conc}
In this paper we have presented \textit{PyHard}, a new analytical tool intended to address a gap in meta-learning studies about instance hardness for a single dataset, distributed publicly at PyPI\footnote{\url{https://pypi.org/project/pyhard/}}. The tool allows for the construction, visualization and interaction with a hardness embedding of a dataset, allowing to obtain useful insights to support data-centric analysis. Our case study demonstrates the utility of the tool for identifying anomalous and noisy observations within a dataset and to highlight observations for which a given classifier shows a consistently good predictive performance and where it may be biased. 

\section*{Acknowledgements}

To the São José dos Campos municipal health secretariat and to the research agencies CAPES - Finance Code 001 (main grant 88887.507037/2020-00), CNPq (grant 307892/2020-4) and the Australian Research Council (grant FL140100012).

{\small
\bibliography{reference}}
\bibliographystyle{apalike}

\end{document}